\documentclass[10pt,twocolumn,letterpaper]{article}

\usepackage[pagenumbers]{cvpr} 

\definecolor{cvprblue}{rgb}{0.21,0.49,0.74}
\usepackage[pagebackref,breaklinks,colorlinks,allcolors=cvprblue]{hyperref}
\usepackage{amsmath,amssymb}

\def\paperID{*****} 
\def\confName{CVPR}
\def\confYear{2026}

\title{Zero-shot 3D General Obstacle Detection via Multimodal Foundation Models and Geometry}

\author{Tamás Matuszka\\
aiMotive\\
{\tt\small tamas.matuszka@aimotive.com}
\and
Péter Hajas\\
aiMotive\\
{\tt\small peter.hajas@aimotive.com}
\and
Dávid Szeghy\\
aiMotive\\
{\tt\small david.szeghy@aimotive.com}
}

\begin{document}
\maketitle
\begin{abstract}

Detecting general obstacles is critical for autonomous driving, especially in long-tail scenarios with rare or unseen objects. Existing methods rely on supervision or predefined categories, limiting generalization. We propose a training-free approach that combines multimodal foundation models with geometric reasoning for 3D obstacle detection. Our key idea is to detect obstacles as deviations from the road surface, segmented in 2D and localized in 3D via temporal LiDAR aggregation. The pipeline operates in a zero-shot manner without task-specific training. Experiments show accurate localization up to 100 meters and 10–25\% recall gains from foundation model priors, while enabling scalable autolabeling.
\end{abstract}    
\section{Introduction}
\label{sec:intro}

Numerous 3D perception algorithms in the literature have been successfully applied to autonomous driving with reliable performance. These algorithms are based on different concepts, such as transformers \cite{li2022bevformer}, \cite{Chen_2023_CVPR}, and convolutional neural networks \cite{liang2022bevfusion}, \cite{liu2023bevfusion}, \cite{philion2020lift}, and they perform well on predefined categories defined by established datasets. However, autonomous vehicles operating in real-world scenarios need to be able to recognize categories beyond these predefined classes to ensure safe and reliable deployment. Since general obstacles may fall into long-tail distribution object classes or categories that were not seen during training, traditional supervised learning methods are not suitable for detecting hazardous obstacles on the road surface. Additionally, incorporating a new obstacle category into the training set would require an expensive retraining phase, which hinders the scalable development of perception algorithms as obstacle classes are not known in advance.

There are several methods available for addressing the problem of general obstacle detection, but these approaches have significant limitations. Existing approaches, such as stereo vision, depth estimation, and semantic segmentation, struggle to reliably detect small, distant, or previously unseen obstacles, especially under real-world conditions.

This raises the question: can general obstacles be detected in 3D without costly training? We address this by combining foundation models with geometric reasoning. Our key insight is that obstacles can be identified as deviations from the road surface, which can be robustly segmented in image space and localized in 3D via temporal aggregation of LiDAR data.

Foundation models, such as \textit{Grounding DINO} \cite{liu2023grounding} and \textit{Segment Anything} \cite{kirillov2023segment}, enable promptable and general-purpose perception through large-scale pretraining. We leverage these models to extract obstacle candidates by focusing on the road surface, while clustering and tracking algorithms are used to achieve consistent 3D localization over time. Unlike open-world detection methods \cite{joseph2021towards}, our objective is not to classify unknown objects but to detect their presence, eliminating the need for retraining or oracle labeling. As a result, the proposed pipeline operates in a training-free, zero-shot manner and generalizes to previously unseen obstacle types in real-world driving scenarios.

In summary, this paper makes the following contributions:
\begin{itemize}
\item A unified approach combining foundation models and geometric reasoning for general obstacle detection.
\item A training-free (zero-shot), offline method for 3D general obstacle detection enabling large-scale autolabeling without manual supervision.

\end{itemize}

\section{Related work}
\label{sec:rel_work}

Open World Detection (OWD) aims to identify objects outside predefined categories and incrementally learn them over time \cite{joseph2021towards}. Existing approaches typically rely on region proposal networks \cite{ren2015faster, joseph2021towards, kim2022learning, wu2022uc, wang2023detecting}, which exhibit bias toward known classes and require supervised training. In contrast, our method is training-free and operates directly in 3D, which is essential for autonomous driving.

3D open-vocabulary methods leverage pretrained multimodal models for tasks such as segmentation, detection, and occupancy prediction \cite{ding2023pla, cao2023coda, vobecky2023pop}. However, these approaches still require training and often struggle with small or distant. Similarly, contrastive learning-based methods \cite{liu2023segment, bhalgat2023contrastive} depend on costly training procedures.

Drivable free space and obstacle detection methods \cite{popov2023nvradarnet, garnett2017real, levi2015stixelnet} rely on manually or semi-automatically generated ground truth, limiting scalability and robustness to rare obstacle types.

Recent multimodal autolabeling approaches \cite{najibi2023unsupervised} reduce annotation effort but still require expensive training (e.g., knowledge distillation) and may miss small obstacles due to ground removal.

In contrast, training-free instance segmentation methods \cite{deng2023elc, takmaz2023openmask3d} rely on pretrained segmentation models that often classify unknown objects as background. Our approach overcomes this limitation through targeted prompting of foundation models combined with geometric reasoning.

\begin{figure}[t]
  \centering
  \includegraphics[width=0.95\linewidth]{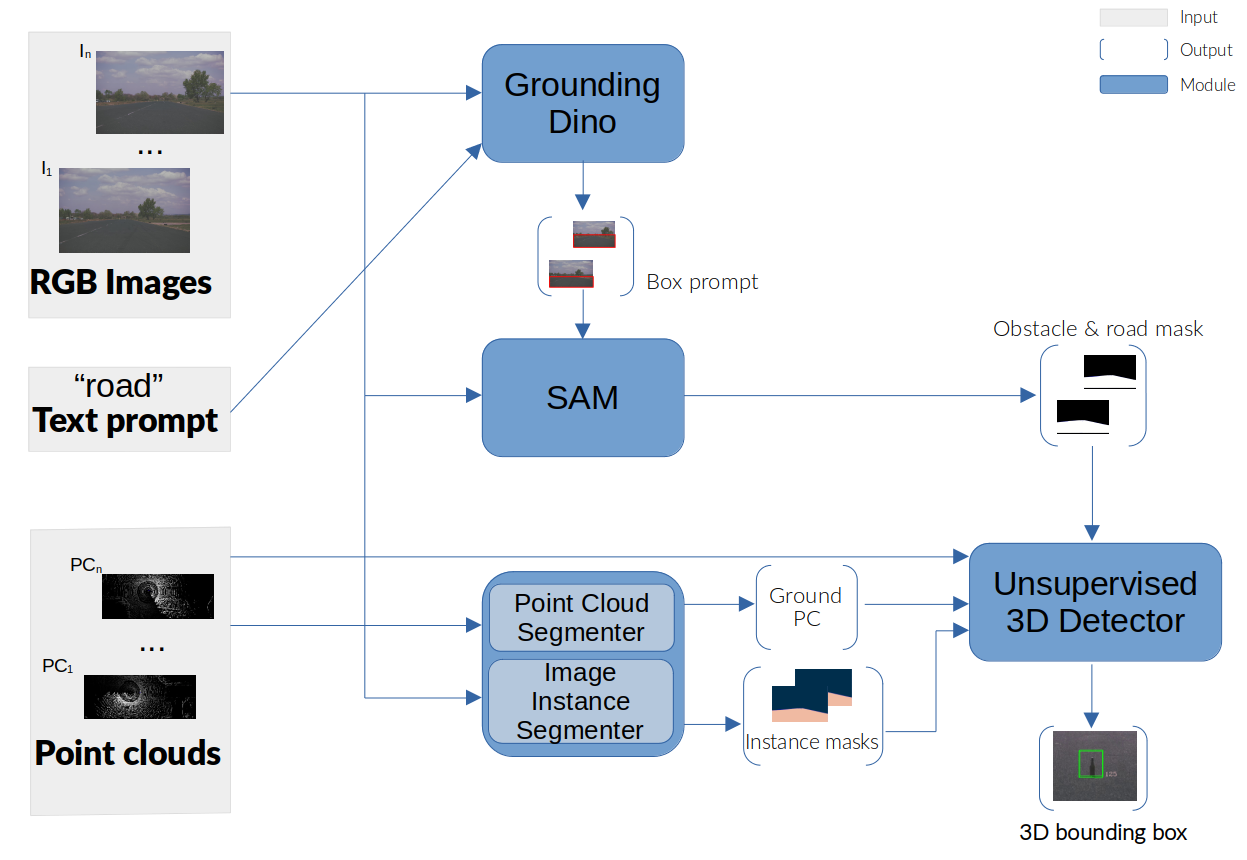}
  \caption{The architecture of the proposed obstacle detection method. The upper side of the figure depicts the foundation model-based general obstacle segmentation, while the lower part is the training-free offline detector that is responsible for localization in 3D space. Best viewed by zooming in.}
  \label{fig:method}
\end{figure}

\section{Hybrid 3D Obstacle Detection}
\subsection{Problem Statement}

We consider the problem of detecting 3D bounding boxes of general obstacles from synchronized LiDAR and camera sequences. Given multimodal inputs and ego-motion, the goal is to localize both known objects and previously unseen obstacles on the drivable road surface.

\begin{figure*}[t]  
  \centering
  \includegraphics[width=1.0\linewidth]{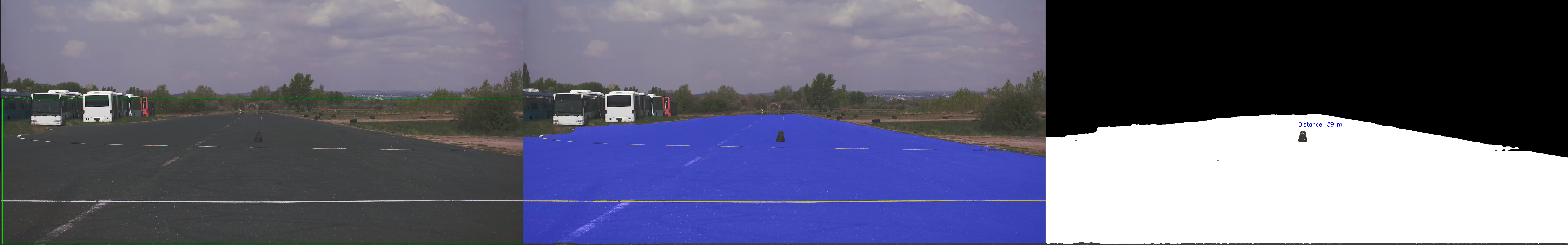}
  \caption{One result of the foundation model-based obstacle segmentation method. Left: bounding box determined by Grounding DINO prompted with the word 'road'. Middle: road mask determined by SAM prompted by the bounding box given by Grounding DINO. Right: obstacle candidate mask with an initial depth estimation from projected LiDAR points.}
  \label{fig:found-simple}
\end{figure*}

\begin{figure*}[t]
    \centering
    \begin{subfigure}{1.0\textwidth}
        \includegraphics[width=\textwidth]{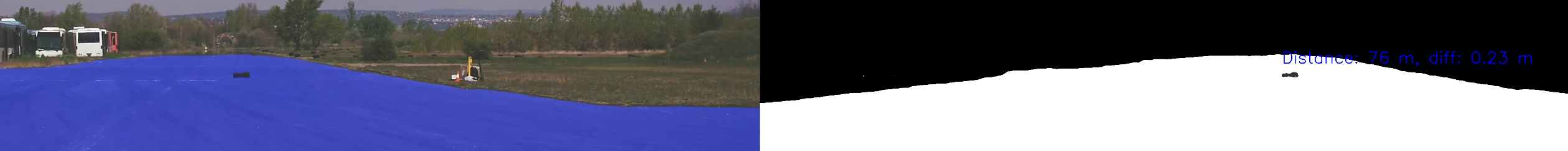}
        \caption{Example of successful depth estimation ($\leq 0.3$ m error) for a tire detected on the road surface.}
        \label{fig:first}
    \end{subfigure}
    \hfill
    \begin{subfigure}{1.0\textwidth}
        \includegraphics[width=\textwidth]{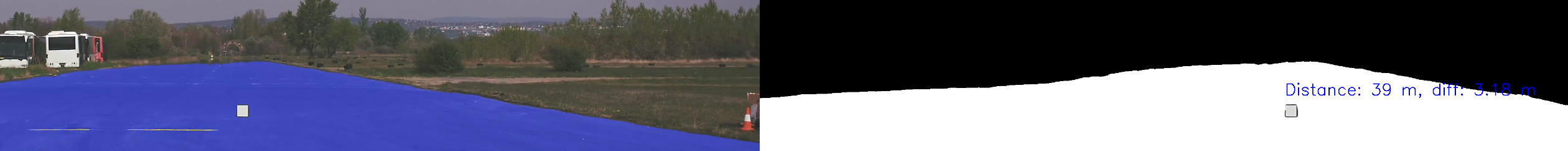}
        \caption{Erroneous depth estimation (3.18 m error) of a white cardboard box due to insufficient LiDAR reflections.}
        \label{fig:second}
    \end{subfigure}            
    \caption{Qualitative results of the naive method.}
    \label{fig:naive}
\end{figure*}

\subsection{Multimodal Foundation Model-based General Obstacle Segmentation}
\label{method:foundational}

The first part of our method uses foundation models to segment general obstacle candidates on a per-frame basis. We reformulate obstacle detection as identifying deviations from the road surface, rather than directly detecting object categories. The upper part of Figure \ref{fig:method} illustrates the pipeline, which resembles GroundedSAM \cite{ren2024grounded}. We refer to elements of the figure using \textit{italic} text. Standard semantic segmentation networks are unsuitable for this task, as they classify unknown objects as background, leading to missed obstacles on the road surface. Open-vocabulary detection is also insufficient due to the high variability of obstacle types, which prevents the use of a fixed category list such as ['debris', 'obstacle', 'stuff'].

Instead, we detect \textit{road surface masks} in \textit{RGB images} and identify their discontinuities as obstacle candidates. We use \textit{Grounding DINO} with the word "road" as a \textit{text prompt} to localize the road surface via bounding boxes, which are then used to \textit{box prompt} the \textit{SAM} module. Due to its extensive pretraining, \textit{SAM} can separate the road surface from small or distant objects, enabling reliable obstacle candidate extraction. A convex hull-based post-processing step ensures completeness of partially covered objects (e.g., near the horizon). While some false positives remain, they are removed in the subsequent stage. Figure \ref{fig:found-simple} shows an example of the proposed segmentation method.


\subsection{Geometry-based 3D Localization}
\label{method:compgeo}

The second part of the proposed method utilizes computational geometry and tracking algorithms and is illustrated by the bottom side of Figure \ref{fig:method}. The key idea is to transform per-frame segmentation cues into consistent 3D obstacle hypotheses through temporal aggregation and geometric consistency. Our approach enhances the 3D bounding boxing method introduced in \cite{matuszka2023aimotive}. The \textit{Unsupervised 3D Detector} involves four primary steps with input obtained from various sources.

(1) The \textit{Point Cloud Segmenter} (provided by e.g., PTv3 \cite{wu2024ptv3}) is used to identify ground points (\textit{Ground PC}) in each frame. After removing the detected ground points, the point cloud is segmented into clusters. Density-based clustering algorithms like DBSCAN \cite{ester1996density} can be used to obtain these point cloud clusters due to their robustness to noise. The clusters are then matched and tracked in consecutive frames, resulting in an aggregated point cloud for each object candidate that appears in the perceived world as seen by the LiDAR from different viewpoints. We utilized the Generalized-ICP algorithm \cite{segal2009generalized} for building an aggregated point cloud for each object candidate. With the assistance of the \textit{instance masks} produced by \textit{Image Instance Segmenter} (we used Mask2Former \cite{cheng2022masked}) and \textit{Point cloud} projection onto \textit{RGB images}, these object candidates are classified into a predefined list of categories. This allows for closed-set object detection to be performed. However, the unknown objects on the road surface with classes not included by the \textit{Image Instance Segmenter}'s class list will not be detected due to the drawbacks of supervised segmentation networks mentioned in \ref{method:foundational}. We also retain these tracked obstacle candidates with unknown categories for further analysis taking into account their position relative to the road, geometric properties (extent), and whether the track has a long enough lifetime. These obstacle candidates will later be used for open-world detection. The points of removed \textit{Ground PC} are also classified as 'road' and 'non-road' clusters using the \textit{Image Instance Segmenter} and \textit{Road mask}.

(2) The \textit{Point clouds} from the LiDAR frames are projected onto the closest image in time after compensating for motion. The LiDAR points that belong to the \textit{obstacle masks} calculated by the foundation models identify potential general obstacle clusters in the point cloud frame that are not detected by the \textit{Image Instance Segmenter}. To reduce the number of false positive detections, clusters with extreme size have been removed from obstacle candidates. Furthermore, inconsistent and unstable detections were also removed from the candidate pool. After this refinement, the remaining clusters are used as input for the matching-tracking method mentioned in (1) to obtain tracked objects.

(3) Using \textit{Ground PC} that was extracted in (1), we construct a world model of the road by incorporating the road points from all the LiDAR frames. Additionally, we can identify clusters on the road surface that are outside the expected distribution using statistical data analysis by processing the points locally and with their neighborhood, fitting a non-linear plane, considering deviations, and thresholding. These clusters represent stationary objects on the road, including bumps and holes, which shall be detected but might not captured by the obstacle segmentation results of the foundation models.

(4) Finally, a post-processing step is required because the tracked objects on the road surface may be detected from various sources (e.g., objects belonging to known categories found by the \textit{Image Instance Segmenter}, and general obstacles proposals determined by \textit{obstacle masks} or method (3)). This post-processing step implements an entity resolution method by combining the detections from different sources and unifying them to handle multiple detections obtained from various methods. As a result, our method generates \textit{3D bounding boxes} for closed-set categories and unknown general obstacles by fitting cuboids on the aggregated point clouds of tracked closed-set objects and general obstacles.
\section{Experiments}

\subsection{General Obstacle Dataset}
\label{eval:dataset}
Existing datasets such as Lost and found \cite{pinggera2016lost} and SegmentMeIfYouCan \cite{chan2021segmentmeifyoucan} lack LiDAR support or sufficient range for 3D obstacle detection. Therefore, we recorded an in-house dataset with 151 sequences (34k frames), including LiDAR and camera data, with manual annotations up to 100 meters.

\begin{table}[]
\begin{tabular}{l|l|ll|ll}
Mode & Long. ROI & \multicolumn{2}{c}{w/o mask} &  \multicolumn{2}{c}{with mask} \\
     &     & Prec & Rec & Prec & Rec \\
\hline
 All & [-100,100]& 94.7 & 58.5 & 93.4  & 69.4 \\
 All & [-50,80] &  94.5&  74.8 & 95.3 & 84.6  \\
 \hline
 LiDAR-vis. & [-100,100]& 94.2 & 68.9 & 96.2 & 89.9 \\
 LiDAR-vis. & [-50,80]& 93.9 & 79.0 & 96.2 & 97.9 \\
 LiDAR-vis. & [10,80]&  95.2 & 74.2 &  97.0 & 99.3 \\
\end{tabular}
\caption{Precision and recall metrics of obstacle detection. All: all frames used, even when the obstacle is not visible by LiDAR, LiDAR-vis: filtered frames where the obstacle is visible by LiDAR. Recall is significantly higher using the masks given by the foundation models.}
\label{tab:quant}
\end{table}

\subsection{Naive baseline method}
The first approach relied solely on the foundation model-based obstacle segmentation described in \ref{method:foundational} without utilizing temporality and point cloud aggregation. In possession of the obstacle candidate masks and the raw LiDAR measurements corresponding to the same frame, a distance estimate can be obtained for each obstacle by averaging the depth component of the projected LiDAR points onto the obstacle masks. This solution provides sensible depth estimates for obstacles with a sufficient amount of reflected LiDAR points. However, the depth estimation error for debris made from non-reflective material is large (in the order of magnitude in meters). The depth cannot even be estimated in the absence of LiDAR reflections. The average longitudinal displacement using the naive method is 1.573 m (std: 1.68 m, max: 7.5 m) when obstacle masks can be associated with LiDAR reflections. Figure \ref{fig:naive} shows an accurate (less than 30 cm localization error) and unsuitable (larger than 1-meter localization error) depth estimation provided by the naive method. Furthermore, the 3D extent cannot be determined by employing the naive approach. These experiments reinforced the need for an offline solution that can rely on non-causality and computational geometry for processing the aggregated point cloud. 

\begin{figure}[t]
    \centering
    \begin{subfigure}{0.49\textwidth}
        \includegraphics[width=\textwidth]{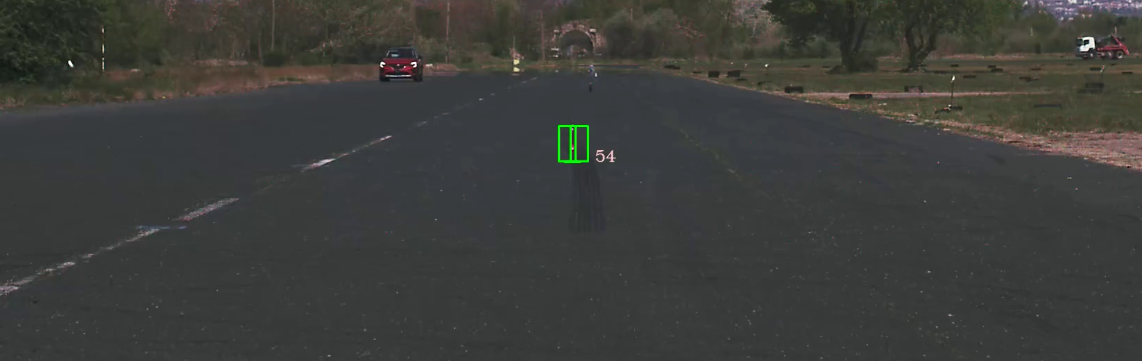}
        \label{fig:succ1}
    \end{subfigure}
    \hfill
    \begin{subfigure}{0.49\textwidth}
        \includegraphics[width=\textwidth]{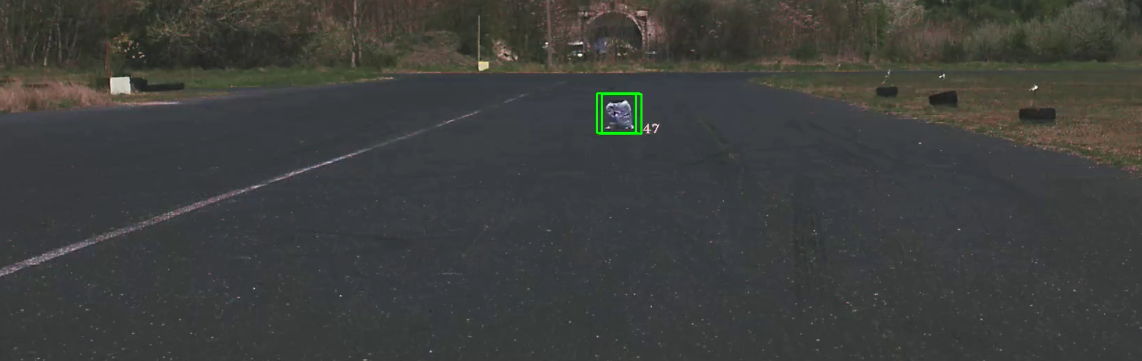}
        \label{fig:succ2}
    \end{subfigure}      
        \begin{subfigure}{0.49\textwidth}
        \includegraphics[width=\textwidth]{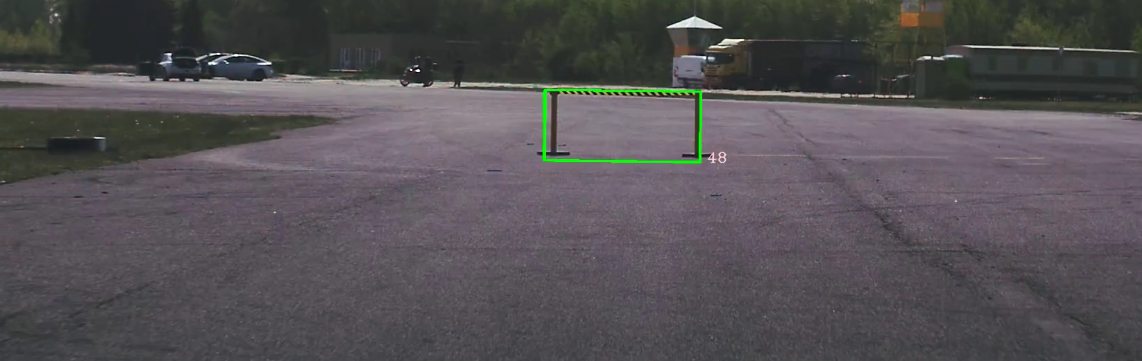}
        \label{fig:succ3}
    \end{subfigure}
    \hfill
    \vspace{-10pt}
    \caption{Successful detection of diverse obstacle types, including small, distant, and non-reflective objects.}
    \label{fig:dets}
\end{figure}

\subsection{Quantitative results}

Table \ref{tab:quant} reports precision and recall for the proposed method with and without the use of obstacle masks. We evaluate localization performance under two settings: (1) all annotated obstacles are considered, including those outside the LiDAR field of view, and (2) only obstacles visible to the LiDAR are retained. In addition, results are analyzed across different perception ranges to assess performance as a function of distance.

The proposed approach is capable of detecting general obstacles up to 100 meters from the ego vehicle. Incorporating obstacle masks derived from multimodal foundation models consistently improves recall by 10–25\% across all evaluated settings, while maintaining comparable precision. This highlights the importance of foundation model-based priors for recovering difficult or partially observed obstacles. When restricting the evaluation to LiDAR-visible obstacles, the method achieves near-perfect performance in the forward-looking direction. 

\noindent\textbf{Comparison to naive baseline.}
The naive per-frame approach, which directly estimates depth from projected LiDAR points, suffers from large localization errors (1.57 m average longitudinal error) and fails when LiDAR reflections are sparse or missing. In contrast, the proposed method leverages temporal aggregation and geometric reasoning to achieve accurate and robust 3D localization, with longitudinal and lateral errors below 0.65 m and 0.57 m,.


The primary limitation arises when obstacles are not observable by the LiDAR sensor, in which case reliable 3D localization is not possible. As a result, recall decreases by approximately 5–10\% when such cases are included, as reflected in Table \ref{tab:quant}. This behavior is expected, as the method fundamentally relies on LiDAR measurements for accurate depth estimation and 3D reconstruction.
\section{Conclusion}


We presented a training-free method for 3D general obstacle detection that combines multimodal foundation models with geometric reasoning. The approach enables the detection of unseen obstacles and achieves accurate localization. Unlike prior methods, it requires no dataset-specific adaptation and is directly applicable across environments, enabling scalable 3D autolabeling and large-scale dataset generation from unlabeled driving data. A limitation is its reliance on LiDAR visibility, preventing reliable localization of occluded obstacles.
{
    \small
    \bibliographystyle{ieeenat_fullname}
    \bibliography{main}
}


\end{document}